# Symbolic Autoencoding for Self-Supervised Sequence Learning

**Mohammad Hossein Amani** [1]  **Nicolas Mario Baldwin** [1]  **Amin Mansouri** [1]  **Martin Josifoski** [1]  **Maxime Peyrard** [2]
**Robert West** [1]

## Abstract

Traditional language models, adept at next-token prediction in text sequences, often struggle with transduction tasks between distinct symbolic systems, particularly when parallel data is scarce. Addressing this issue, we introduce *symbolic autoencoding* (ΣAE), a self-supervised framework that harnesses the power of abundant unparallel data alongside limited parallel data. ΣAE connects two generative models via a discrete bottleneck layer and is optimized end-to-end by minimizing reconstruction loss (simultaneously with supervised loss for the parallel data), such that the sequence generated by the discrete bottleneck can be read out as the transduced input sequence. We also develop gradient-based methods allowing for efficient self-supervised sequence learning despite the discreteness of the bottleneck. Our results demonstrate that ΣAE significantly enhances performance on transduction tasks, even with minimal parallel data, offering a promising solution for weakly supervised learning scenarios.

## 1. Introduction

The field of artificial intelligence has undergone a remarkable transformation in recent years, propelled by the rise of powerful language models. At the heart of this success lie sequence-to-sequence (seq2seq) transducers, a class of models trained to transform sequences of symbols or tokens from one language system to another. For instance, mapping from one natural language to another (machine translation), from text to structured representations (part-of-speech tagging, semantic role labeling, etc.), and maintaining discourse in natural language through chatbots.

The core challenge for these models lies in inferring the mapping $M$ between the two language systems, denoted as $X$ and $Z$, for a specific task, such that $Z = M(X)$. Recent large language models display striking emergent abilities to perform many such mappings after exposure to massive but diverse textual data. However, they fail when one or both language systems are scarce or nonexistent in the training data, or when the mapping function deviates significantly from patterns present during training ([Magueresse et al.,](#) [2020;](#) [Lample & Conneau, 2019;](#) [Joshi et al., 2020](#)). To still learn the mapping $M$, the conventional strategy involves fine-tuning a pre-trained language model using the limited parallel data available. If sufficient parallel data is available, a more direct approach is to train the model from scratch.

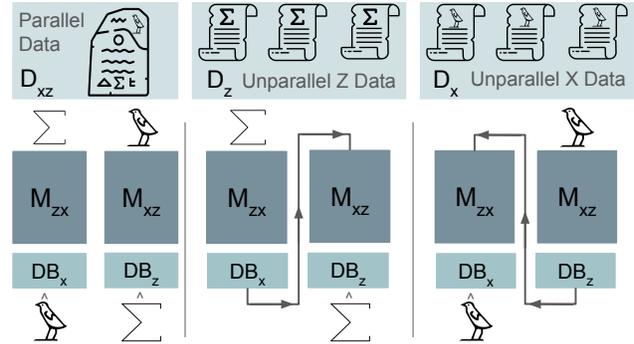

*Figure 1.* Illustration of the abstract flow of data in the symbolic autoencoding (ΣAE) framework, exemplified with the Rosetta Stone problem. Two sequence-to-sequence models ($M_{xz}$ and $M_{zx}$) are trained with both parallel data (the Rosetta Stone) through next-token prediction and unparallel data through connecting the models with a discrete bottleneck layer ($DB_x$ and $DB_z$) to autoencode each language using the other as its hidden representation.

In this work, we explore an alternative paradigm that breaks away from the dependency on pre-trained models. We introduce a novel method that effectively leverages small amounts of parallel data, even if insufficient for standard supervised training or fine-tuning. Compared to similar previous works that focused on specific tasks such as machine translation, summarization, or non-sequential data, our method is generally applicable to any type of sequence-to-sequence modeling, including structured or formal languages.

---

[1]Department of Computer Science, EPFL, Lausanne, Switzerland [2]Univ. Grenoble Alpes, CNRS, Grenoble INP, LIG. Correspondence to: Mohammad Hossein Amani <mh.amani1998@gmail.com>.

Our setup of interest finds a natural parallel in the historical challenge of deciphering the ancient Egyptian hieroglyphs from the Rosetta stone (see Fig. 1). Even if Egyptian hieroglyphs were abundant in ancient Egyptian papyri, they remained a mystery until the discovery of the Rosetta Stone in 1799, which provided the key to unlocking the secrets of the hieroglyphs (Budge, 1913). The stone slab bears the same text inscribed in three distinct scripts: Egyptian hieroglyphs, Demotic script, and ancient Greek. This limited parallel dataset, a mere 27 lines of text, was sufficient to guide researchers in understanding the full translation $M$ between ancient Egyptian hieroglyphs ($X$) and ancient Greek ($Z$).

To address the machine learning generalization of the Rosetta Stone problem, we introduce *symbolic autoencoding* ($\Sigma$AE). $\Sigma$AE simultaneously learns the mapping $M_{xz}$ from $X$ to $Z$ and the mapping $M_{zx}$ from $Z$ to $X$, in a symmetrical manner. This is achieved by orchestrating several losses that reuse $M_{xz}$ and $M_{zx}$ with varying input and output data. First, two supervised losses are used to tune $M_{xz}$ and $M_{zx}$ on the scarce parallel data. Due to the scarcity of parallel data, these alone are not sufficient to learn the mappings. Therefore, we introduce two additional autoencoding losses. One autoencoder reconstructs $z \in Z$ after encoding it discretely into $x \in X$ via the path $Z \xrightarrow{M_{zx}} X \xrightarrow{M_{xz}} Z$, where $M_{zx}$ is the decoder and $M_{xz}$ is the encoder. The connection between the two is implemented by a *discrete bottleneck*, a novel differentiable mechanism that binds two sequence-to-sequence networks into an end-to-end, fully differentiable sequence-to-sequence-to-sequence model. The discrete bottleneck serves as the differentiable glue between $M_{xz}$ and $M_{zx}$. Symmetrically, the second autoencoder reconstructs $x$ after encoding it into $z$.

These two autoencoding losses utilize the larger sets of non-parallel data, explicitly enforcing the constraint of always employing the mappings $M_{xz}$ and $M_{zx}$. This treats $X$ as the hidden representation for the auto-encoding of $Z$, and vice versa. The supervised losses further ensure that not any arbitrary languages are learned as the hidden representations.

Our main contributions are the following:

- We introduce $\Sigma$AE as a framework for connecting two seq2seq models via discrete bottlenecks (DB) and training them using gradient-based optimization.
- We propose and evaluate techniques to improve the efficiency of sequence learning within the $\Sigma$AE framework, including various DB implementations.
- We empirically demonstrate the superior performance of $\Sigma$AE over traditional supervised baselines using both synthetic and real-world data.

To facilitate further research, we open-source code and data.

**Related work.** Our work is related to several directions in unsupervised and weakly supervised learning through discrete representations. Baziotis et al. (2019) connected two encoder–decoder models via a hidden sequence layer, utilizing a reconstruction loss and a language model prior loss for unsupervised sentence compression. Kaiser & Bengio (2018) explored semantic hashing (Salakhutdinov & Hinton, 2009) and the Gumbel softmax trick (Jang et al., 2017) for creating interpretable, discrete encodings. Similar approaches to discretization, including the use of continuous paths alongside discrete ones, were examined by Fortuin et al. (2019), reflecting the challenges of training with discrete bottlenecks. Kingma et al. (2014) studied semi-supervised autoencoding by treating the classification label as a categorical hidden variable. Moreover, several researchers have studied discretization of elements and representations in neural networks (Liu et al., 2022; 2021; Tamkin et al., 2023; Peng et al., 2018; Maddison et al., 2016). Our approach stands apart by (i) employing straightforward negative log-likelihood loss for efficient reconstruction without relying on auxiliary losses, (ii) enabling discretization across sequences of variable lengths, (iii) blending both unsupervised and supervised training data in a novel way, and (iv) integrating sequence models as building blocks for creating sequential autoencoders.

## 2. Preliminaries

### 2.1. Notation

We consider a seq2seq modeling task, with the goal of learning mappings $M_{xz}$ and $M_{zx}$ from a sequence of discrete symbols $x \in X$ to a sequence $z \in Z$ and vice versa. Here $X = V_x^*$ and $Z = V_z^*$ are the Kleene closures of the finite set of symbols $V_x$ and $V_z$, respectively. The vocabulary sizes of $X$ and $Z$ are denoted by $|V_x|$ and $|V_z|$.

We denote the $t$-th element of the sequence $x$ by $x^t$, and of $z$ by $z^t$. The lengths of these sequences are denoted by $T_x$ and $T_z$. Elements $x^t \in V_x$ and $z^t \in V_z$ are tokens, which means they could be words, characters, or sub-words, depending on the nature of the sequences and the choice of tokenization in forming the task.

We assume we are given a dataset comprising three subsets: parallel sequences $D_{xz}$ and unparallel $X$ and $Z$ sequences $D_x$ and $D_z$.

Throughout, vectors are denoted by bold lowercase letters, such as $\mathbf{v}$ and $\mathbf{s}$, while a dictionary of vectors is represented by $D$, without any subscripts. To indicate a specific element $i$ within a dictionary or vector, we append $[i]$ to its symbol. In certain contexts, $i$ may refer to a token; in these instances, $[i]$ refers to the integer ID corresponding to that token.

**Sequence Models.**
In this work, we demonstrate the $\Sigma$AE framework using autoregressive models. Note, however, that it is versatile enough to be applied to single-input/single-output models as well, i.e., models that generate an output sequence in one step ($z = M_{xz}(x)$), as opposed to autoregressive models where the output $z^t$ at position $t$ is dependent on previously generated outputs $z^{<t}$, ($z^t = M_{xz}(x, z^{<t})$).

The transducer models generate output vectors

$$\mathbf{v}_z^t = M_{xz}(x, z^{<t}), \qquad (1)$$
$$\mathbf{v}_x^t = M_{zx}(z, x^{<t}), \qquad (2)$$

where $\mathbf{v}_z^t$ and $\mathbf{v}_x^t$ refer to the output vectors at time step $t$ of models $M_{xz}$ and $M_{zx}$, respectively. With slight abuse of notation, we allow $M_{xz}$ and $M_{zx}$ to also be directly applied to token embeddings rather than discrete tokens.

The transducer model can vary widely, including recurrent neural networks and transformers. This range ensures broad applicability across different data types and task requirements. For example, if transformer encoder–decoder models are used as $M_{xz}$ and $M_{zx}$, then $\mathbf{v}_z^t$ and $\mathbf{v}_x^t$ could be the decoder hidden states. For a decoder-only transformer, they could be the decoder hidden states at position $T_x + t$, with the inputs $x$ and $z^{<t}$ concatenated and fed into the model.

## 2.2. Surrogate Gradients for Discrete Layers

When using a non-differentiable layer, we need to provide surrogate gradient estimations to keep our models learnable with gradient descent. Assume $f_0 : \mathbb{R}^m \to \mathbb{R}^n$ is a non-differentiable function and $f_\tau : \mathbb{R}^m \to \mathbb{R}^n$ a continuous approximation of that function. Then, to avoid directly designing the gradient substitute, we can use the gradient of the approximate function $\nabla f_\tau$ in the backward pass while still applying $f_0$ in the forward computation. An example of such a function pair is $f_0 = \arg\max$ and $f_\tau = \mathrm{softmax}$.

Implementing this gradient substitution in an automatic differentiation library also becomes straightforward. In the computation graph we just need to replace the assignment $x \leftarrow f_0(x)$ with $x \leftarrow f_0(x) + f_\tau(x) - \mathrm{sg}(f_\tau(x))$. Where sg denotes the stop gradient operator, which is the identity function in the forward pass but has zero gradient in the backward pass. This way of passing the gradients has been used in previous works on learning discrete representations (van den Oord et al., 2017; Bengio et al., 2013).

## 3. $\Sigma$AE Framework

We now describe the $\Sigma$AE framework in detail. For an overview, see Fig. 2. In Sec. 3.1 we provide an abstract definition of the discrete bottleneck and explore three distinct implementations of this concept. In Sec. 3.2 we describe a notable challenge in using autoregressive models, the "early EOS problem," where the hidden sequence collapses prematurely. To address this, we propose a feedback mechanism that mitigates this issue, making the sequential autoencoding feasible. Finally, in Sec. 3.3, we introduce training strategies that effectively combine parallel and non-parallel training modes.

## 3.1. Discrete Bottleneck

The discrete bottleneck (DB) acts as a conduit for discrete information flow between models while preserving differentiability. It allows two seq2seq models to be trained simultaneously in an end-to-end manner.

Functionally, the DB provides two outputs:

- **Probability vector s**: It assigns a discrete distribution over tokens, facilitating training with negative log-likelihood loss.
- **Quantized vector $\mathbf{v}_q$**: This vector serves as an input for subsequent models or layers, such as the decoder transducer model in the reconstruction task.

The discreteness of DB implies that the quantized vector should belong to a finite discrete domain $\mathbf{v}_q \in D$. In other words, $D$ should be a finite set of embedding vectors, e.g. a dictionary of concepts or tokens, enforcing the vectors passing through DB to carry at most $\log_2 |D|$ bits of information.

This discrete nature of the DB is what differentiates our paradigm from continuous autoencoding. By confining the model output to a finite set of vectors, we force the model to represent input features as a composition of the dictionary vectors rather than a single continuous vector (Liu et al., 2022). This discrete computation in the DB is also the point of non-differentiability; gradient estimation techniques, as discussed in Sec. 2.2, are employed here to enable the backward propagation of gradients.

### 3.1.1. Training Models with DB Head

In this work we append a DB layer to each seq2seq model, $DB_x$ to $M_{zx}$ and $DB_z$ to $M_{xz}$. This setup allows us to specify (i) separate and (ii) joint training modes (see below). We note that, although the DB dictionaries can have any size, when the DB is appended as the last layer to a model it is intuitive to match its dictionary size with the models' vocabulary sizes, considering we assign one vector to each token.

  i For parallel training data $(x, z) \in D_{xz}$, to predict the $t$-th token $z^t$, the models generate a probability vector over all the tokens in the vocabulary $\mathbf{s}_z^t$ given the input sequence $x$ and previous label tokens $z^{<t}$ (similarly

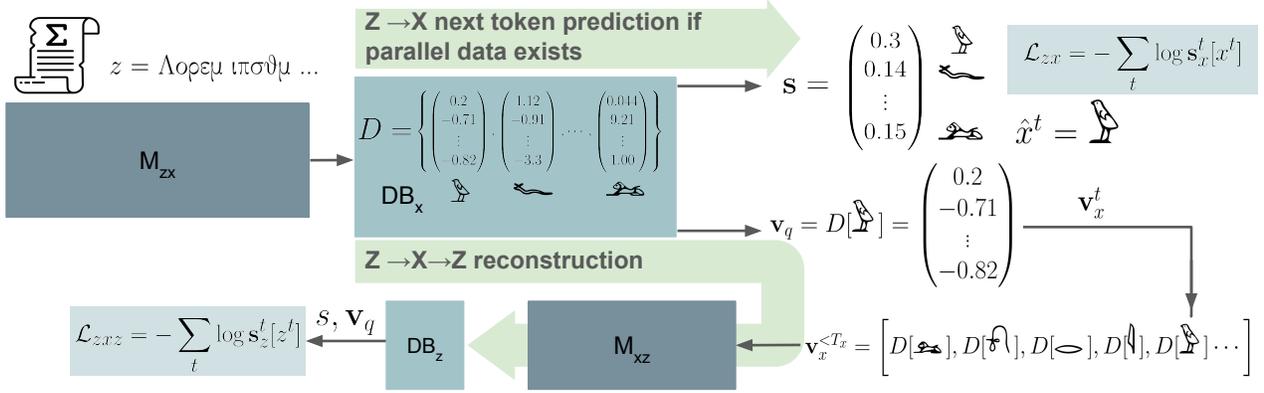

*Figure 2.* **Overview of symbolic autoencoding (ΣAE),** illustrated on the Rosetta Stone problem (see Fig. 1). The discrete bottleneck (DB) generates two outputs: a score vector $\mathbf{s}$ and a quantized vector $\mathbf{v}_q$. When supervised labels $x^t$ are present, $M_{zx}$ is updated via negative log-likelihood loss $\mathcal{L}_{zx} = -\log(\mathbf{s}[x^t])$ on scores $\mathbf{s}$. Even without labels, the forward pass can continue through the unsupervised $z \to x \to z$ reconstruction path. The sequence of DB quantized vectors $\mathbf{v}_x^{<T_x}$ serve as input to $M_{xz}$ to reconstruct the original $z$ (the greek text in the figure). Subsequently, both models, $M_{zx}$ and $M_{xz}$, are updated jointly using the reconstruction loss $\mathcal{L}_{zxz}$, calculated as negative log-likelihood on the decoder's ($M_{xz}$) output scores.

for predicting the $x$ sequence):

$$\mathbf{s}_z^t, \mathbf{v}_z^t = \mathrm{DB}_z(M_{xz}(x, z^{<t})), \quad \mathcal{L}_{xz} = -\sum_t \log \mathbf{s}_z^t[z^t] \tag{3}$$

$$\mathbf{s}_x^t, \mathbf{v}_x^t = \mathrm{DB}_x(M_{zx}(z, x^{<t})), \quad \mathcal{L}_{zx} = -\sum_t \log \mathbf{s}_x^t[x^t] \tag{4}$$

and the seq2seq models are trained separately to minimize their respective negative log-likelihood losses.

ii With unlabeled data $x \in D_x$ (or $z \in D_z$), the models generate a latent sequence of quantized vectors $(\mathbf{v}_x^{<T_x} = \{\mathbf{v}_x^t\}_{t=0}^{T_x})$:

$$\mathbf{s}_x^t, \mathbf{v}_x^t = \mathrm{DB}_x(M_{zx}(z, \mathbf{v}_x^{<t})) \quad \text{or} \tag{5}$$

$$\mathbf{s}_x^t, \mathbf{v}_x^t = \mathrm{DB}_z(M_{xz}(x, \mathbf{v}_z^{<t})). \tag{6}$$

The sequence of quantized vectors are then passed as input to the other model (the decoder) to reconstruct the original input $x$ (or $z$) and train both models end-to-end:

$$\mathbf{s}_z^t, \mathbf{v}_z^t = \mathrm{DB}_z(M_{xz}(\mathbf{v}_x^{<T_x}, z^{<t})) \quad \text{or} \tag{7}$$

$$\mathbf{s}_x^t, \mathbf{v}_x^t = \mathrm{DB}_x(M_{zx}(\mathbf{v}_z^{<T_z}, x^{<t})), \tag{8}$$

using as the reconstruction loss, respectively, $\mathcal{L}_{zxz} = -\sum_t \log \mathbf{s}_z^t[z^t]$ or $\mathcal{L}_{xzx} = -\sum_t \log \mathbf{s}_x^t[x^t]$.

### 3.1.2. Discrete bottleneck implementations

DBs in the ΣAE framework can be broadly classified into two types: **probability-based** and **embedding-based**.

**Probability-based DB.** In a probability-based DB the score function $\mathbf{s} = S(\mathbf{v})$ does not depend on the embedding vectors in the dictionary $D$ and it precedes the realization of the quantized vector $\mathbf{v}_q = V_q(\mathbf{s})$ in the computational graph. This means that computing $\mathbf{v}_q$ requires a decoding step from the distribution $\mathbf{s}$, i.e.

$$\mathbf{s} = S(\mathbf{v}), \quad \mathbf{v}_q = D[\text{decode}(\mathbf{s})]$$

In this work we focus on the softmax function as $S$ and experiment with two decoding methods:

- **Softmax DB** uses maximum likelihood decoding, i.e., the quantized vector $\mathbf{v}_q$ corresponds to the most likely token in the dictionary:

$$\mathbf{s}[i] = \frac{\exp(\mathbf{v}[i])}{\sum_{j=1}^{|V|} \exp(\mathbf{v}[j])} \tag{9}$$

$$\mathbf{v}_q = D\left[\arg\max_i \mathbf{s}[i]\right] \tag{10}$$

- **Gumbel DB** uses categorical sampling for decoding:

$$\mathbf{s}[i] = \frac{\exp(\mathbf{v}[i] + g_i)}{\sum_{j=1}^{|V|} \exp(\mathbf{v}[j] + g_j)} \tag{11}$$

$$\mathbf{v}_q = D\left[\arg\max_i \mathbf{s}[i]\right] \tag{12}$$

Here $g_i$ is a sample from the Gumbel distribution, i.e. $g_i = -\log(-\log(u_i))$ where $u_i \sim \mathrm{Uniform}(0, 1)$. We use the Gumbel reparameterization trick to translate the sampling to taking the argmax of noisy probabilities (Jang et al., 2017).

Crucially, during the backward pass, gradients are passed to $\mathbf{s}$ as if $\mathbf{v}_q$ was computed as the soft average of dictionary embeddings, instead of via argmax or categorical sampling.

This is expressed by assigning $\mathbf{v}_q \leftarrow \mathbf{v}_q + \sum_{i=1}^{|V|} \mathbf{s}[i]D[i] - \text{sg}(\sum_{i=1}^{|V|} \mathbf{s}[i]D[i])$ in an automatic differentiation library.

**Embedding-based DB.** In the embedding-based DB the scores are a function of the dictionary embeddings. As an example of an embedding-based DB, we propose vector-quantized DB (**VQ DB**), which is similar in implementation to VQ-VAE (van den Oord et al., 2017):

$$\mathbf{l}[i] = \|\mathbf{v} - D[i]\| \tag{13}$$

$$\mathbf{v}_q = D\left[\arg\min_i \mathbf{l}[i]\right] \tag{14}$$

$$\mathbf{s}[i] = S(-\mathbf{l}[i]) \tag{15}$$

In the above equations, $\mathbf{l}$ denotes the vector of distances between the input vector $\mathbf{v}$ and the embedding vectors $D[i]$ in the dictionary, and the score is a function of that. Once again we focus on softmax as the probability function $S$ in our experiments. Since $\mathbf{v}_q$ is the closest embedding vector to the input $\mathbf{v}$, we directly pass the gradients from $\mathbf{v}_q$ to $\mathbf{v}$. This is achieved by assigning $\mathbf{v}_q \leftarrow \mathbf{v}_q + \mathbf{v} - \text{sg}(\mathbf{v})$ in the computation graph in an automatic differentiation program.

### 3.1.3. REMARKS

As training progresses, it is notable that models gain confidence in their predictions, leading to more polarized score distributions. This polarization results in the models identifying the most likely token with increasing certainty, making the scores sparser, which in turn improves the accuracy of gradient approximations. Nevertheless, to ensure effective symbolic representation learning, it is critical that models communicate solely through discrete symbols. This strict adherence to discreteness helps avoid the interpretability issues and learning impediments associated with soft quantization in traditional VAEs. By maintaining this level of strictness, the information flow remains within the set bit limit, essential for upholding both the interpretability and the integrity of compositional learning within the $\Sigma$AE framework.

### 3.2. Addressing Hidden Sequence Collapse in Seq2Seq Models

Training seq2seq models with unsupervised data often requires autoregressive generation of hidden tokens until an End-of-Sequence (EOS) token or a maximum length is reached. For non-autoregressive models, the equivalent would be padding out any token after the first time the EOS token was predicted. This process involves a discrete decision about when to halt generation or from where to throw away the rest of the sequence - a decision that we hope the model learns implicitly rather than through explicit gradient-based training. Specifically, the model isn't directly informed that generating an EOS token would stop the generation or filter out the rest of the generated sequence.

### 3.2.1. THE CHALLENGE: FIRST TOKEN RELIANCE

A prevalent issue in sequential auto-encoders, often exacerbated in the context of a discrete bottleneck, is the tendency to overly rely on the first token of the hidden representation, sidelining the potential of subsequent tokens. This problem, documented before (Havrylov & Titov, 2020; Newman et al., 2020), leads to underutilization of the hidden sequence and suboptimal local minima. Without explicit feedback, the model remains unaware of its ability to control the EOS, often encoding all necessary information into the first few tokens.

A potential justification of this phenomena could be the following: Without providing explicit feedback, the effect of assigning a high probability to the EOS token remains unknown to the model. In other words, from the point of view of the model, information packed after a stochastic position in the sentence are dropped out. Unaware that it can control the EOS, the model is then motivated to encode all information in only the first token of the hidden representation, to learn the most robust way to pass the information about the input features to the decoder. This effect resembles how models learn more robust representations in presence of dropout (Srivastava et al., 2014) by encoding information in fewer units.

### 3.2.2. EOS MASKING WITH GRADIENT APPROXIMATION

In unsupervised training scenarios, halting points can vary across different samples in a batch. To address this, usually generation continues until either the maximum length is reached or all samples have produced an EOS token. Subsequently, tokens generated post-EOS are masked out. The same masking procedure is used for single-I/O models.

Assuming the generation of $T$ tokens, with the first EOS token at $T_{<\text{EOS}>} < T$, we define a binary mask $\mathbf{m}$ of size $T$ where each $\mathbf{m}[i]$ is 1 if the EOS token has not been generated and 0 otherwise:

$$\mathbf{m}[i] = \begin{cases} 1 & \text{if } \mathbf{m}[i-1] = 1 \text{ and } O_{i-1} \neq \text{<EOS>} \\ 0 & \text{otherwise} \end{cases} \tag{16}$$

In this configuration, $O$ represents the output sequence generated by a model, either a sequence in $X$ or $Z$. Applying the mask $M$ to the quantized vectors $\mathbf{v}_q^{<T}$ during the forward pass effectively enforces a halting mechanism, setting vectors post-EOS appearance to zero and thereby terminating the sequence generation.

Hence, $\mathbf{m}$ is binary random vector of the form

$$P(\mathbf{m}[i] = 1) =$$
$$\left(1 - \mathbb{P}(O_{(i-1)} = \text{<EOS>})\right) \mathbb{P}(\mathbf{m}[i-1] = 1)$$
$$= \prod_{k=1}^{i-1} \left(1 - \mathbb{P}(O_k = \text{<EOS>})\right). \quad (17)$$

Therefore the expected value of $\mathbf{m}$ is

$$\mathbb{E}[\mathbf{m}[i]] = \prod_{k=1}^{i-1} \left(1 - \mathbb{P}(O_k = \text{<EOS>})\right) \quad (18)$$

To counter the problem of autoregressive collapse, in the backward computation we pass the gradients through $\mathbf{m}$ to $\mathbb{P}(O_k = \text{<EOS>})$ as if $\mathbb{E}[\mathbf{m}]$ had been the masking matrix in the forward computation. Thus we provide direct feedback on the EOS effect through another gradient approximation:

$$\mathbf{m} \approx \mathbf{m} + \mathbb{E}[\mathbf{m}] - \text{sg}(\mathbb{E}[\mathbf{m}]) \quad (19)$$

We note that in our experiments, without using this approximation, the unsupervised training almost always failed because of the hidden state collapse. Once again, as training progresses, the models become more confident in predicting EOS correctly, resulting in more polarized probabilities. This means that the expected mask $\mathbb{E}[\mathbf{m}]$ better approximates the true mask, making our approximation more accurate.

### 3.3. Training with Scheduling Strategies

The $\Sigma$AE framework accommodates three training modes due to its access to multiple data sources, as illustrated in Fig. 1. These modes include: (i) **Supervised Training**, using parallel data $D_{xz}$ to minimize cross-entropy losses $\mathcal{L}_{xz}$ and $\mathcal{L}_{zx}$; (ii) **X Reconstruction**, using unparallel data $D_x$ to minimize reconstruction loss $\mathcal{L}_{xzx}$; (iii) **Z Reconstruction**, using unparallel data $D_z$ to minimize reconstruction loss $\mathcal{L}_{zxz}$.

Our framework is adaptable to additional models and data sources. For instance, one can imagine $M_{zy}, M_{yx}, \cdots$ each with their own supervised and reconstruction losses, e.g. $\mathcal{L}_{zy}, \mathcal{L}_{yxz}, \mathcal{L}_{xyx}, \cdots$, to be optimized. However, even with two models and three data sources, the task of simultaneously minimizing these four losses poses a significant challenge. The complexity arises from the intricate interdependencies that exist between the losses with shared models – a characteristic feature of multi-task learning environments. However, unlike certain multi-task scenarios where individual tasks may appear independent or unrelated, in the $\Sigma$AE framework improvement in one task can directly benefit others, creating a synergy that enhances overall performance.

To navigate this multi-objective optimization problem, we propose three scheduling strategies for training the $\Sigma$AE framework:

- Joint Training: At each iteration, we randomly select a batch from $D_{xz}$, $D_x$, or $D_z$, akin to flipping a three-sided coin, and train for one iteration on the corresponding training mode.
- Unsupervised Pretraining with Supervised Finetuning: We initially train on $D_x$ and $D_z$ until convergence, followed by finetuning on $D_{xz}$.
- Supervised Pretraining with Unsupervised Finetuning: We start by training on $D_{xz}$ until convergence, then shift to finetuning on $D_x$ and $D_z$.

In both finetuning approaches, we employ a gradual curriculum shift rather than an abrupt change. This involves slowly altering the probability distribution of the 'three-sided coin', used for batch selection in joint training, to transition smoothly from the initial training phase to the subsequent finetuning phase.

## 4. Experimental Setup

### 4.1. Tasks

For our experiments, we utilized four seq2seq datasets (examples in Table 1): SCAN, PCFG SET, CFQ, and COGS, chosen for their controlled environments and precise accuracy measures:

- SCAN (Lake & Baroni, 2017) is a simple language-driven navigation instruction task designed to evaluate the ability of neural models to learn compositional commands.
- PCFG SET (Hupkes et al., 2019) is a synthetic dataset generated using probabilistic context-free grammars, aimed at testing the systematic generalization of models.
- CFQ (Keysers et al., 2019) is a large-scale dataset of complex natural language questions and their corresponding SPARQL query against the Freebase knowledge base designed to measure the compositional generalization capabilities of semantic parsing models, with questions constructed to reflect the compositional structure of Freebase.
- COGS (Kim & Linzen, 2020): COGS is a dataset for evaluating the generalization of semantic parsing models to novel linguistic structures, emphasizing the model's ability to generalize from given sentences to new sentences that have similar syntactic structures but different lexical items or phrasal constructions.

This selection of datasets ensures a comprehensive and nuanced evaluation of the $\Sigma$AE framework. The datasets also facilitate direct evaluation of our approach, avoiding re-liance on proxy metrics often used in larger datasets. Here, the mapping from $X$ to $Z$ is unique and non-reversible, with $Z$ typically being the longer sequence, serving as a reliable ground truth for $X$. Our study diverges from the typical use of these datasets for compositional generalization; instead of focusing on out-of-distribution testing, we emphasize in-distribution performance assessment. We also conduct a bidirectional evaluation of both $M_{xz}$ and $M_{zx}$ models, reflecting realistic seq2seq model applications where translation in both directions holds equal significance, in line with the suggestions of Bastings et al. (2018).

*Table 1.* Example of Samples from Different Datasets

| Dataset | Sample |
|---|---|
| SCAN | **X:** look right thrice after run left <br> **Z:** I_TURN_LEFT I_RUN I_TURN_RIGHT I_LOOK <br> I_TURN_RIGHT I_LOOK I_TURN_RIGHT I_LOOK |
| PCFG SET | **X:** echo append append E18 C13 , <br> L18 M17 , R1 L1 Y1 T18 J18 <br> **Z:** E18 C13 L18 M17 R1 L1 Y1 T18 J18 J18 |
| CFQ | **X:** Who influenced M1 's cinematographer , writer , and editor <br> **Z:** SELECT DISTINCT ?x0 WHERE <br> ?x0 a ns:people.person. <br> ?x0 ns:influence.influence_node.influenced ?x1. <br> ?x1 ns:film.cinematographer.film M1. <br> ?x1 ns:film.editor.film M1. <br> ?x1 ns:film.writer.film M1. |
| COGS | **X:** Olivia rolled Liam. <br> **Z:** roll . agent ( x_1 , Olivia ) AND roll . theme ( x_1 , Liam ) |

### 4.2. Experimental Tasks in $\Sigma$AE Framework

To demonstrate the effectiveness of the $\Sigma$AE framework, we conducted two sets of experiments on each of the datasets. These experiments are designed to showcase the framework's capabilities under different training conditions:

- **Unsupervised Training**: In this scenario, we assume access only to unparallel data, with the primary goal being to reconstruct $Z$ from a hidden discrete sequence. The framework is configured such that the dictionary size and the maximum sequence length of the hidden representation are matched to those of $X$. This setup tests the $\Sigma$AE framework's ability to autonomously compress the input and perform accurate reconstructions.

- **Weakly-supervised training**: This scenario simulates a situation akin to the Rosetta Stone problem, where a small portion of the data is parallel and the rest is unparallel. The objective here is to incorporate unsupervised data, aiming to minimize both the unsupervised losses ($\mathcal{L}_{zxz}$ and $\mathcal{L}_{xzx}$) and supervised losses ($\mathcal{L}_{zx}$ and $\mathcal{L}_{xz}$). We conduct four different experiments for each dataset and DB implementation, varying the supervision ratio $\eta = \frac{|D_{xz}|}{|D_{xz}|+|D_x|+|D_z|}$ defined as the portion of the training set that is parallel. This allows us to assess the framework's efficiency in leveraging limited paral-

lel data to enhance performance on larger unparallel datasets.

### 4.3. Model Architecture and Hyperparameters

In our experiments with the $\Sigma$AE framework, we adopted a standardized model architecture and hyperparameter setting across all tasks to maintain consistency and focus on the framework's effectiveness. We utilized a six-layer transformer encoder–decoder model for $M_{xz}$ and $M_{zx}$. We used greedy decoding consistently for all tasks, simplifying the decoding process and ensuring uniformity across experiments. Model learning rates were manually chosen on the order of $10^{-3}$ or $10^{-4}$, to ensure a decrease in loss during the early stages of training. Hyperparameters were not extensively tuned. For each task, the same hyperparameters were used across different supervision ratios. This uniform approach underscores the framework's robustness, although we acknowledge that more nuanced tuning and regularization might yield higher performance.

### 4.4. Metrics

In assessing the performance of the $\Sigma$AE framework, we measured two distinct metrics: **sentence accuracy (SA)** and **token accuracy (TA)**. These metrics are designed to provide both a holistic and a detailed view of the model's capabilities. Sentence accuracy (SA) for a sample is counted as 1 if the entire sentence is correctly generated. Token accuracy (TA) is a more granular measure, where correctness of each predicted token in all sentences are measured separately. This metric allows for partial credit within a sentence, providing a finer understanding of the model's performance at the token level.

The token accuracy can be measured with two methods: We can *teacher-force* the correct previous tokens (as per the ground truth) to the model and measure its accuracy in predicting the next token. Alternatively, the model's previous outputs (which may or may not be correct) can be used as inputs for generating subsequent tokens. This *autoregressive* approach is generally more challenging than teacher-forcing.

Each $X$ has a unique corresponding $Z$, simplifying the assessment of accuracy in this direction, therefore, evaluating $M_{xz}$ performance is simply done by examining the *Autoregressive Z TA/SA*, directly measuring the model's capability to generate accurate $Z$ sequences. For a given $Z$, however, there could be multiple valid $X$ sequences. Therefore, to evaluate $M_{zx}$, we utilize the *Teacher-forced X TA*, which restricts the range of correct $X$ sequences for end tokens. Another approach is the *Reconstruction Z TA/SA*, where a model $M_{xz}$ maps a generated sequence $\hat{x}$ back to $Z$, and the accuracy of this reconstructed sequence serves as a proxy for the correctness of $\hat{x}$.

## 5. Experimental Results

Our investigation into the $\Sigma$AE framework, spanning four distinct datasets and five metrics, showcases significant achievements in unsupervised reconstruction (detailed in Table 2) and weakly supervised training tasks (Fig. 3), with an emphasis on the benchmarks of Softmax DB and a specific focus on $Z$ sentence accuracy. Given the uniformity of results across various settings, we've compiled extensive plots for the Gumbel and VQ discretizers in the appendix for a more comprehensive analysis.

**Unsupervised Training.** In the unsupervised scenario, the Softmax Discrete Bottleneck (DB) emerged as a particularly effective discretizer, achieving over 98% token accuracy on SCAN, CFQ, and COGS datasets, though performance dipped on the PCFG SET dataset. This variation in results led us to hypothesize that the unique symbolic nature of variables within the PCFG SET task—where basic tokenization assigns distinct representations to symbolically equivalent variables—may underlie the observed performance discrepancy. The Gumbel and VQ DBs also performed similarly, leading us to believe that the models have been effectively trained in all scenarios.

**Weakly Supervised Training.** In the $Z$ space, the Softmax DB consistently surpassed supervised baselines, significantly enhancing token and sentence accuracy across all datasets. For instance, with only 8% supervision on the PCFG SET dataset, token accuracy improved from below 15% to above 80%. While the Gumbel DB generally showed noisier training and slightly weaker performance, it still outperformed supervised baselines in most scenarios, except for a minor shortfall in the COGS dataset at a 16% supervision ratio. The VQ DB, despite showing a slight weaker performance in supervised baselines, improved the training similar to the Softmax and Gumbel DBs, achieving over 20% token accuracy on CFQ dataset at 2% supervision ratio.

While no single Discrete Bottleneck or scheduling method universally outperforms others across all datasets and supervision ratios, for every dataset and $\eta$ value, at least one of our scheduling methods consistently surpasses the baseline performance. In other words, training within the $S\Sigma$AE paradigm always enhances performance, though the optimal choice of the scheduling strategy depends on the task.

The $\Sigma$AE framework's impact extends into the $X$ space, where the Softmax, Gumbel, and VQ DBs exhibit performance boosts. Notably, the exception to this trend occurs with teacher-forced token accuracy in the SCAN dataset for the Softmax DB, indicating a unique challenge in this specific setting.

We note that the VQ DB faced a peculiar issue of numerical instability on the SCAN dataset after extended training periods (+500 epochs). This instability was addressed through weight clipping, suggesting that while $\Sigma$AE offers substantial benefits, optimizing stability and accuracy across different data representations and tasks may require tailored adjustments. These insights into performance variations across $X$ and $Z$ spaces not only highlight the framework's broad applicability but also pinpoint areas for future refinement to maximize the $\Sigma$AE framework's effectiveness.

For all our experiments, we computed 95% confidence intervals via bootstrapped resampling of the test set, however they are too small to be visible on the plots. This performance analysis underscores the $\Sigma$AE framework's versatility and its capacity to leverage both unsupervised and weakly supervised data to enhance model training and performance across diverse seq2seq tasks.

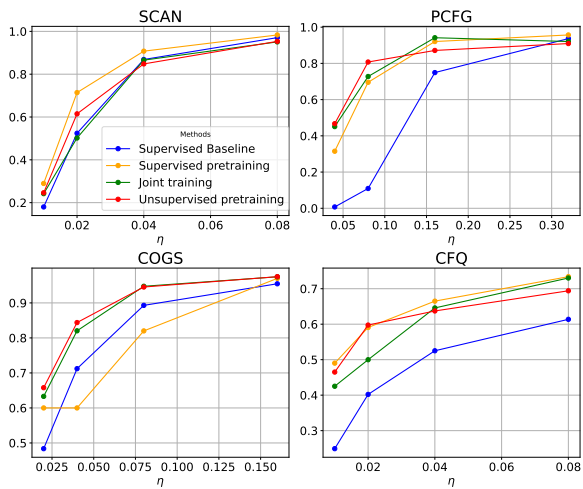

*Figure 3.* Results for Softmax Discrete bottleneck – $Z$ Autoregressive Sentence Accuracy per Supervision Ratio ($\eta$). The blue line shows the accuracy of a model trained only on supervised data. At each $\eta$, one of our scheduling methods from Sec. 3.3 outperforms this supervised baseline. The accuracy of models improve with more supervised data, therefore the performance gaps tightens as the accuracies converge to their maxima. The rest of the performance metrics described in Sec. 4.4 are presented in Fig. 4 for softmax DB, Fig. 5 for Gumbel DB, and Fig. 6 for VQ DB.

## 6. Discussion

The exploration of autoencoding for sequences, especially in weakly-supervised settings, reveals significant challenges and opportunities for advancement. Our approach, while effective in reducing unsupervised reconstruction loss, does not always translate directly to improved model perfor-mance. A key limitation is the slower training pace of autoregressive models, which cannot be parallelized due to the sequential nature of generating tokens.

*Table 2.* Table of Test Autoregressive token accuracy ($Z$) (top) and Sentence Accuracy ($Z$) (bottom) on the unsupervised task ($Z$ reconstruction)

|  | SCAN | PCFG | COGS | CFQ |
|---|---|---|---|---|
| Softmax DB | 1.00 | 0.74 | 0.98 | 0.99 |
|  | 0.96 | 0.31 | 0.55 | 0.69 |
| Gumbel DB | 0.98 | 0.75 | 0.98 | 0.99 |
|  | 0.74 | 0.36 | 0.51 | 0.43 |
| VQ DB | 1.00 | 0.44 | 0.94 | 0.90 |
|  | 0.93 | 0.00 | 0.03 | 0.00 |

Future directions for this research include enhancing sample efficiency, as our findings suggest unsupervised samples are less impactful on accuracy compared to supervised ones. Additionally, adopting a variational output approach and employing the reparametrization trick could align our framework with the VAE framework, using the ELBO loss for potentially more effective training. The $\Sigma$AE framework's adaptability to various seq2seq models and its applicability across different data modalities—from text to images, audio, or video—highlight its broad utility. This flexibility suggests numerous pathways for further exploration and application beyond the current study's scope.

**Conclusion.** In this study, we introduced a novel approach for training sequence-to-sequence models using non-parallel data. This was achieved by connecting two models through a discrete bottleneck, allowing the output sequence from one model to serve as input for the other, creating a unique autoencoder architecture where both the encoder and decoder are seq2seq models targeted for training. To ensure the end-to-end trainability of this autoencoder within a gradient descent framework, we proposed gradient substitute and autoregressive masking techniques. Our unsupervised experiments validated the feasibility of training models within this discrete sequential autoencoder setup.

Further expanding on this concept, we developed the $\Sigma$AE framework, leveraging the symbolic autoencoders to train seq2seq models beyond parallel data and facilitate the use of non-parallel data. Demonstrating a practical application of our methodology, we presented evidence of performance improvements in weak supervision settings by utilizing unsupervised monolingual data, tested across four distinct seq2seq datasets.

## Acknowledgments

Robert West's lab is partly supported by grants from Swiss National Science Foundation (200021_185043, TMSGI2_211379), H2020 (952215), Microsoft Swiss Joint Research Center, and by generous gifts from Facebook, Google, and Microsoft.

## Impact Statements

This paper presents work whose goal is to advance the field of machine learning. We do not see major specific societal consequences that go beyond what applies to machine learning in general.

## References

Bastings, J., Baroni, M., Weston, J., Cho, K., and Kiela, D. Jump to better conclusions: Scan both left and right. *ArXiv*, abs/1809.04640, 2018. URL https://api.semanticscholar.org/CorpusID:52273813.

Baziotis, C., Androutsopoulos, I., Konstas, I., and Potamianos, A. Seq^3: Differentiable sequence-to-sequence-to-sequence autoencoder for unsupervised abstractive sentence compression. In *North American Chapter of the Association for Computational Linguistics*, 2019. URL https://api.semanticscholar.org/CorpusID:102352497.

Bengio, Y., Léonard, N., and Courville, A. C. Estimating or propagating gradients through stochastic neurons for conditional computation. *ArXiv*, abs/1308.3432, 2013. URL https://api.semanticscholar.org/CorpusID:18406556.

Budge, S. E. A. W. *The Rosetta stone*. London: British Museum, 1913.

Fortuin, V., Hüser, M., Locatello, F., Strathmann, H., and Rätsch, G. Som-vae: Interpretable discrete representation learning on time series. In *International Conference on Learning Representations*, 2019. URL https://api.semanticscholar.org/CorpusID:56657849.

Havrylov, S. and Titov, I. Preventing posterior collapse with levenshtein variational autoencoder. *CoRR*, abs/2004.14758, 2020. URL https://arxiv.org/abs/2004.14758.

Hupkes, D., Dankers, V., Mul, M., and Bruni, E. Compositionality decomposed: How do neural networks generalise? *J. Artif. Intell. Res.*, 67:757–795, 2019. URL https://api.semanticscholar.org/CorpusID:211259383.

Jang, E., Gu, S., and Poole, B. Categorical reparameterization with gumbel-softmax. In *5th International Conference on Learning Representations, ICLR 2017, Toulon, France, April 24-26, 2017, Conference Track Proceedings*. OpenReview.net, 2017. URL https://openreview.net/forum?id=rkE3y85ee.

Joshi, P., Santy, S., Budhiraja, A., Bali, K., and Choudhury, M. The state and fate of linguistic diversity and inclusion in the NLP world. In Jurafsky, D., Chai, J., Schluter, N., and Tetreault, J. (eds.), *Proceedings of the 58th Annual Meeting of the Association for Computational Linguistics*, pp. 6282–6293, Online, July 2020. Association for Computational Linguistics. doi: 10.18653/v1/2020.acl-main.560. URL https://aclanthology.org/2020.acl-main.560.

Kaiser, L. and Bengio, S. Discrete autoencoders for sequence models. *ArXiv*, abs/1801.09797, 2018. URL https://api.semanticscholar.org/CorpusID:34455434.

Keysers, D., Schärli, N., Scales, N., Buisman, H., Furrer, D., Kashubin, S., Momchev, N., Sinopalnikov, D., Stafiniak, L., Tihon, T., Tsarkov, D., Wang, X., van Zee, M., and Bousquet, O. Measuring compositional generalization: A comprehensive method on realistic data. *ArXiv*, abs/1912.09713, 2019. URL https://api.semanticscholar.org/CorpusID:209439843.

Kim, N. and Linzen, T. Cogs: A compositional generalization challenge based on semantic interpretation. *ArXiv*, abs/2010.05465, 2020. URL https://api.semanticscholar.org/CorpusID:222290851.

Kingma, D. P., Mohamed, S., Jimenez Rezende, D., and Welling, M. Semi-supervised learning with deep generative models. In Ghahramani, Z., Welling, M., Cortes, C., Lawrence, N., and Weinberger, K. (eds.), *Advances in Neural Information Processing Systems*, volume 27. Curran Associates, Inc., 2014. URL https://proceedings.neurips.cc/paper_files/paper/2014/file/d523773c6b194f37b938d340d5d02232-Paper.pdf.

Lake, B. M. and Baroni, M. Generalization without systematicity: On the compositional skills of sequence-to-sequence recurrent networks. In *International Conference on Machine Learning*, 2017. URL https://api.semanticscholar.org/CorpusID:46761158.

Lample, G. and Conneau, A. Cross-lingual language model pretraining. *ArXiv*, abs/1901.07291, 2019.

Liu, D., Lamb, A., Kawaguchi, K., Goyal, A., Sun, C., Mozer, M. C., and Bengio, Y. Discrete-valued neural communication. In *Neural Information Processing Systems*, 2021. URL https://api.semanticscholar.org/CorpusID:235743226.

Liu, D., Lamb, A., Ji, X., Notsawo, P. J. T., Mozer, M. C., Bengio, Y., and Kawaguchi, K. Adaptive discrete communication bottlenecks with dynamic vector quantization. *ArXiv*, abs/2202.01334, 2022. URL https://api.semanticscholar.org/CorpusID:246485419.

Maddison, C. J., Mnih, A., and Teh, Y. W. The concrete distribution: A continuous relaxation of discrete random variables. *ArXiv*, abs/1611.00712, 2016. URL https://api.semanticscholar.org/CorpusID:14307651.

Magueresse, A., Carles, V., and Heetderks, E. Low-resource languages: A review of past work and future challenges. *ArXiv*, abs/2006.07264, 2020.

Newman, B., Hewitt, J., Liang, P., and Manning, C. D. The EOS decision and length extrapolation. *CoRR*, abs/2010.07174, 2020. URL https://arxiv.org/abs/2010.07174.

Peng, H., Thomson, S., and Smith, N. A. Backpropagating through structured argmax using a spigot. *ArXiv*, abs/1805.04658, 2018. URL https://api.semanticscholar.org/CorpusID:44143441.

Salakhutdinov, R. and Hinton, G. E. Semantic hashing. *Int. J. Approx. Reason.*, 50:969–978, 2009. URL https://api.semanticscholar.org/CorpusID:1501682.

Srivastava, N., Hinton, G. E., Krizhevsky, A., Sutskever, I., and Salakhutdinov, R. Dropout: a simple way to prevent neural networks from overfitting. *J. Mach. Learn. Res.*, 15:1929–1958, 2014. URL https://api.semanticscholar.org/CorpusID:6844431.

Tamkin, A., Taufeeque, M., and Goodman, N. D. Codebook features: Sparse and discrete interpretability for neural networks. *ArXiv*, abs/2310.17230, 2023. URL https://api.semanticscholar.org/CorpusID:264490402.

van den Oord, A., Vinyals, O., and Kavukcuoglu, K. Neural discrete representation learning. In Guyon, I., von Luxburg, U., Bengio, S., Wallach, H. M., Fergus, R., Vishwanathan, S. V. N., and Garnett, R. (eds.), *Advances in Neural Information Processing Systems 30: Annual Conference on Neural Information Processing Systems 2017, December 4-9, 2017, Long Beach, CA, USA*, pp.

6306–6315, 2017. URL https://proceedings.neurips.cc/paper/2017/hash/7a98af17e63a0ac09ce2e96d03992fbc-Abstract.html.

## A. Appendix

Standard errors in all tables are less than 0.01, and thus omitted for better readability.

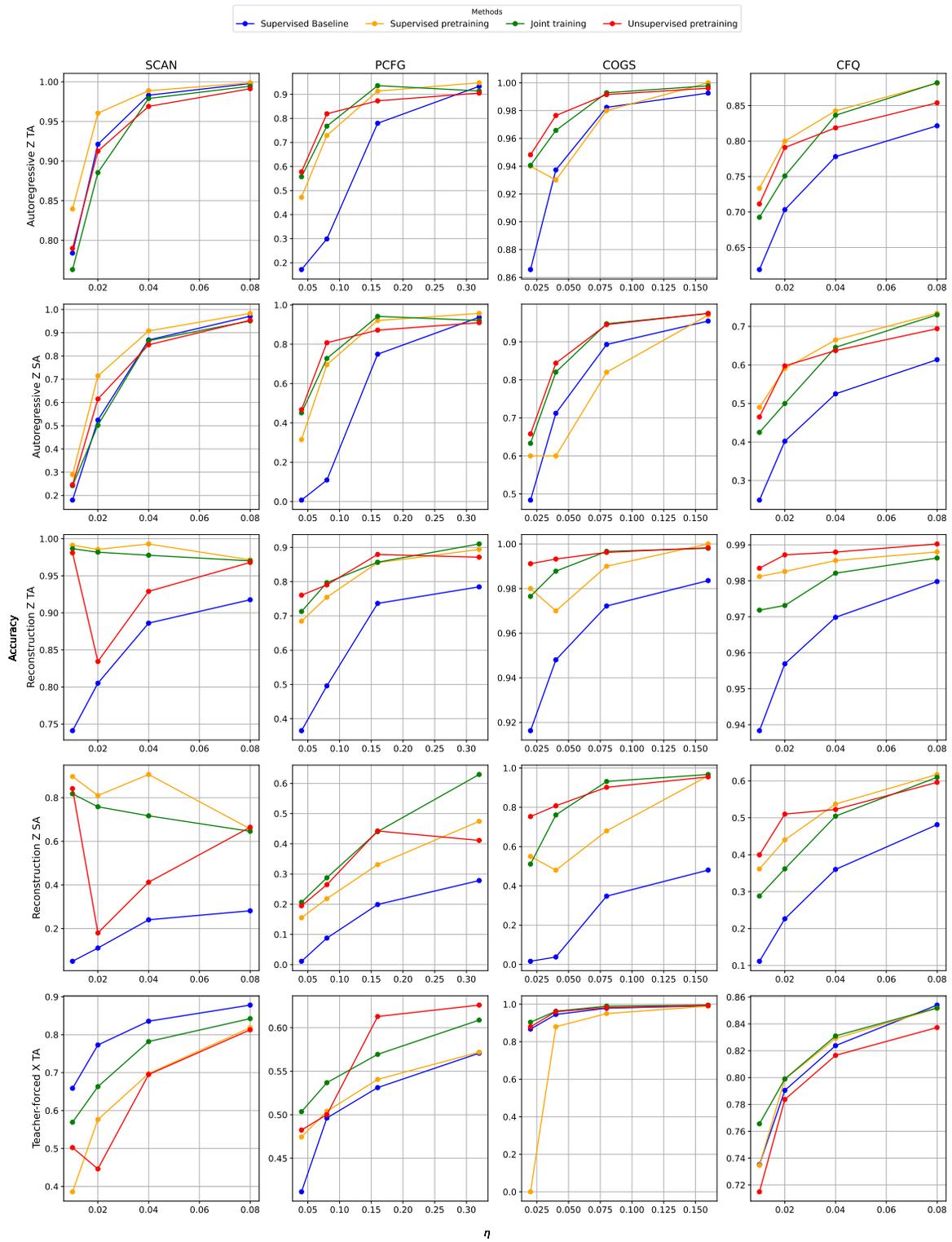

*Figure 4.* Softmax DB performance metrics

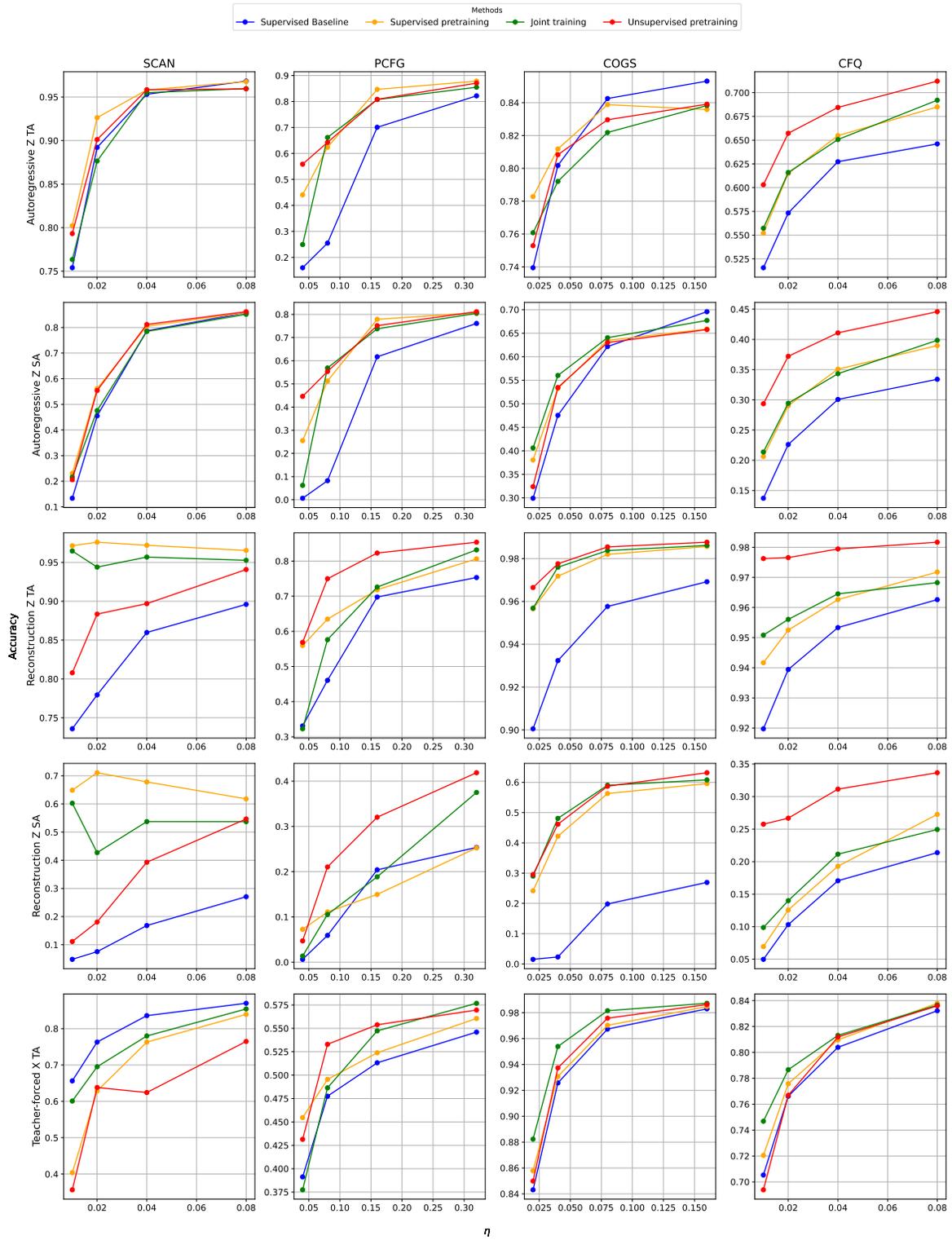

*Figure 5.* Gumbel DB performance metrics

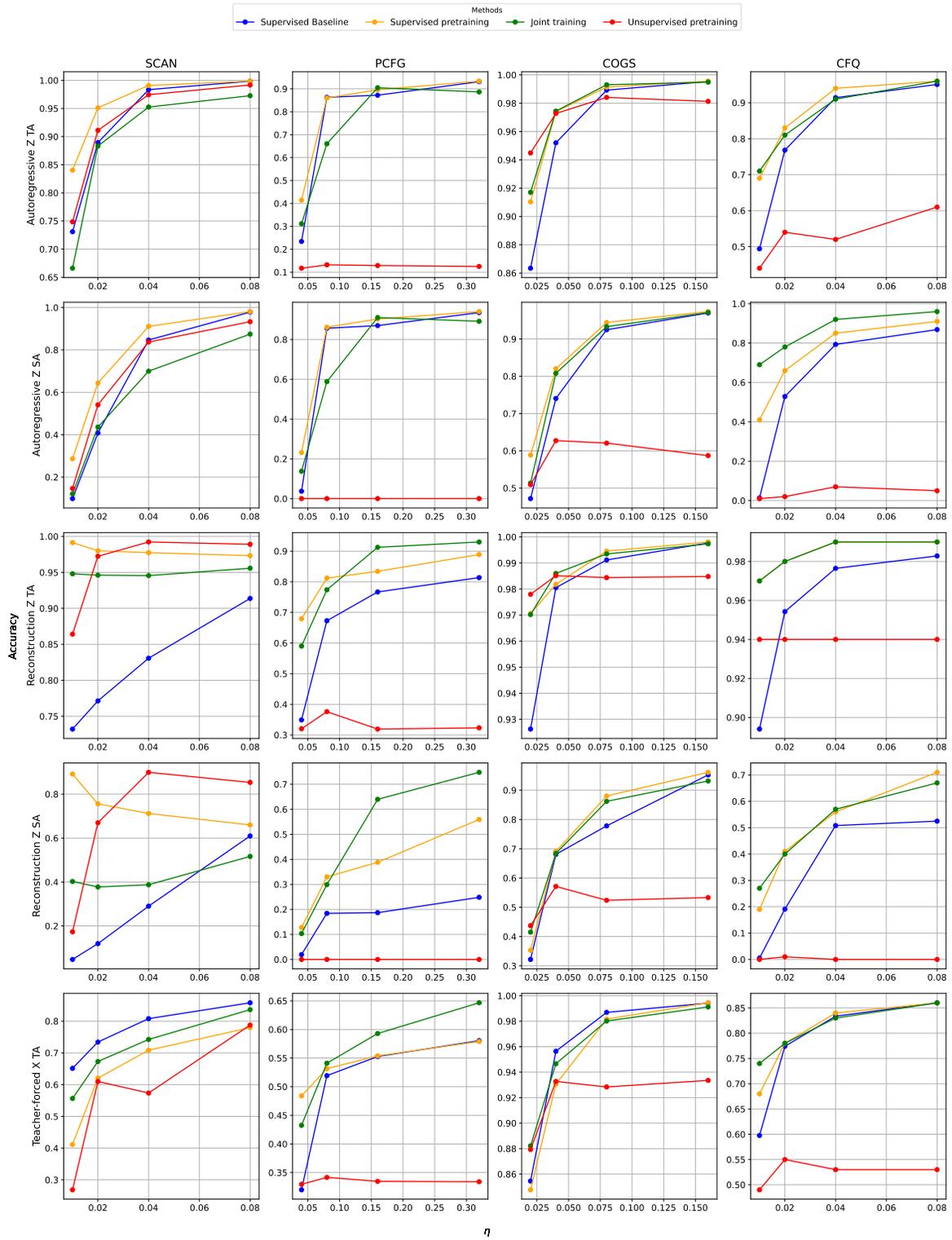

*Figure 6.* VQ DB performance metrics

*Table 3.* Softmax DB – Autoregressive Z TA.

| SCAN | $\eta = 0.04$ | $\eta = 0.08$ | $\eta = 0.16$ | $\eta = 0.32$ | $\eta = 0.99$ |
|---|---|---|---|---|---|
| Supervised Baseline | 0.78 | 0.92 | 0.98 | 1.00 | 1.00 |
| Joint training | 0.76 | 0.89 | 0.98 | 0.99 | — |
| Supervised Pretraining | 0.84 | 0.96 | 0.99 | 1.00 | — |
| Unsupervised Pretraining | 0.79 | 0.91 | 0.97 | 0.99 | — |
| PCFG | $\eta = 0.04$ | $\eta = 0.08$ | $\eta = 0.16$ | $\eta = 0.32$ | $\eta = 0.99$ |
| Supervised Baseline | 0.17 | 0.30 | 0.78 | 0.93 | 0.97 |
| Joint training | 0.56 | 0.77 | 0.94 | 0.91 | — |
| Supervised Pretraining | 0.47 | 0.73 | 0.91 | 0.95 | — |
| Unsupervised Pretraining | 0.58 | 0.82 | 0.87 | 0.91 | — |
| COGS | $\eta = 0.02$ | $\eta = 0.04$ | $\eta = 0.08$ | $\eta = 0.16$ | $\eta = 0.99$ |
| Supervised Baseline | 0.87 | 0.94 | 0.98 | 0.99 | 1.00 |
| Joint training | 0.94 | 0.97 | 0.99 | 1.00 | — |
| Supervised Pretraining | 0.94 | 0.93 | 0.98 | 1.00 | — |
| Unsupervised Pretraining | 0.95 | 0.98 | 0.99 | 1.00 | — |
| CFQ | $\eta = 0.01$ | $\eta = 0.02$ | $\eta = 0.04$ | $\eta = 0.08$ | $\eta = 0.99$ |
| Supervised Baseline | 0.62 | 0.70 | 0.78 | 0.82 | 0.86 |
| Joint training | 0.69 | 0.75 | 0.84 | 0.88 | — |
| Supervised Pretraining | 0.73 | 0.80 | 0.84 | 0.88 | — |
| Unsupervised Pretraining | 0.71 | 0.79 | 0.82 | 0.85 | — |

*Table 4.* Softmax DB – Autoregressive Z SA

| SCAN | $\eta = 0.04$ | $\eta = 0.08$ | $\eta = 0.16$ | $\eta = 0.32$ | $\eta = 0.99$ |
|---|---|---|---|---|---|
| Supervised Baseline | 0.18 | 0.52 | 0.87 | 0.97 | 1.00 |
| Joint training | 0.24 | 0.50 | 0.87 | 0.95 | — |
| Supervised Pretraining | 0.29 | 0.71 | 0.91 | 0.98 | — |
| Unsupervised Pretraining | 0.25 | 0.61 | 0.85 | 0.95 | — |
| PCFG | $\eta = 0.04$ | $\eta = 0.08$ | $\eta = 0.16$ | $\eta = 0.32$ | $\eta = 0.99$ |
| Supervised Baseline | 0.01 | 0.11 | 0.75 | 0.94 | 0.97 |
| Joint training | 0.45 | 0.73 | 0.94 | 0.92 | — |
| Supervised Pretraining | 0.32 | 0.70 | 0.92 | 0.96 | — |
| Unsupervised Pretraining | 0.47 | 0.81 | 0.87 | 0.91 | — |
| COGS | $\eta = 0.02$ | $\eta = 0.04$ | $\eta = 0.08$ | $\eta = 0.16$ | $\eta = 0.99$ |
| Supervised Baseline | 0.48 | 0.71 | 0.89 | 0.95 | 1.00 |
| Joint training | 0.63 | 0.82 | 0.95 | 0.97 | — |
| Supervised Pretraining | 0.60 | 0.83 | 0.94 | 0.97 | — |
| Unsupervised Pretraining | 0.66 | 0.84 | 0.95 | 0.97 | — |
| CFQ | $\eta = 0.01$ | $\eta = 0.02$ | $\eta = 0.04$ | $\eta = 0.08$ | $\eta = 0.99$ |
| Supervised Baseline | 0.25 | 0.40 | 0.53 | 0.61 | 0.69 |
| Joint training | 0.43 | 0.50 | 0.65 | 0.73 | — |
| Supervised Pretraining | 0.49 | 0.59 | 0.66 | 0.73 | — |
| Unsupervised Pretraining | 0.47 | 0.60 | 0.64 | 0.69 | — |

*Table 5.* Softmax DB – Reconstruction Z TA

| SCAN | $\eta = 0.04$ | $\eta = 0.08$ | $\eta = 0.16$ | $\eta = 0.32$ | $\eta = 0.99$ |
|---|---|---|---|---|---|
| Supervised Baseline | 0.74 | 0.81 | 0.89 | 0.92 | 0.96 |
| Joint training | 0.99 | 0.98 | 0.98 | 0.97 | — |
| Supervised Pretraining | 0.99 | 0.99 | 0.99 | 0.97 | — |
| Unsupervised Pretraining | 0.98 | 0.83 | 0.93 | 0.97 | — |
| **PCFG** | $\eta = 0.04$ | $\eta = 0.08$ | $\eta = 0.16$ | $\eta = 0.32$ | $\eta = 0.99$ |
| Supervised Baseline | 0.37 | 0.50 | 0.74 | 0.78 | 0.83 |
| Joint training | 0.71 | 0.80 | 0.86 | 0.91 | — |
| Supervised Pretraining | 0.68 | 0.75 | 0.86 | 0.89 | — |
| Unsupervised Pretraining | 0.76 | 0.79 | 0.88 | 0.87 | — |
| **COGS** | $\eta = 0.02$ | $\eta = 0.04$ | $\eta = 0.08$ | $\eta = 0.16$ | $\eta = 0.99$ |
| Supervised Baseline | 0.92 | 0.95 | 0.97 | 0.98 | 0.99 |
| Joint training | 0.98 | 0.99 | 1.00 | 1.00 | — |
| Supervised Pretraining | 0.98 | 0.97 | 0.99 | 1.00 | — |
| Unsupervised Pretraining | 0.99 | 0.99 | 1.00 | 1.00 | — |
| **CFQ** | $\eta = 0.01$ | $\eta = 0.02$ | $\eta = 0.04$ | $\eta = 0.08$ | $\eta = 0.99$ |
| Supervised Baseline | 0.94 | 0.96 | 0.97 | 0.98 | 0.99 |
| Joint training | 0.97 | 0.97 | 0.98 | 0.99 | — |
| Supervised Pretraining | 0.98 | 0.98 | 0.99 | 0.99 | — |
| Unsupervised Pretraining | 0.98 | 0.99 | 0.99 | 0.99 | — |

*Table 6.* Softmax DB – Reconstruction Z SA

| SCAN | $\eta = 0.04$ | $\eta = 0.08$ | $\eta = 0.16$ | $\eta = 0.32$ | $\eta = 0.99$ |
|---|---|---|---|---|---|
| Supervised Baseline | 0.05 | 0.11 | 0.24 | 0.28 | 0.46 |
| Joint training | 0.82 | 0.76 | 0.72 | 0.65 | — |
| Supervised Pretraining | 0.90 | 0.81 | 0.91 | 0.66 | — |
| Unsupervised Pretraining | 0.84 | 0.18 | 0.41 | 0.66 | — |
| **PCFG** | $\eta = 0.04$ | $\eta = 0.08$ | $\eta = 0.16$ | $\eta = 0.32$ | $\eta = 0.99$ |
| Supervised Baseline | 0.01 | 0.09 | 0.20 | 0.28 | 0.35 |
| Joint training | 0.21 | 0.29 | 0.44 | 0.63 | — |
| Supervised Pretraining | 0.15 | 0.22 | 0.33 | 0.47 | — |
| Unsupervised Pretraining | 0.19 | 0.26 | 0.44 | 0.41 | — |
| **COGS** | $\eta = 0.02$ | $\eta = 0.04$ | $\eta = 0.08$ | $\eta = 0.16$ | $\eta = 0.99$ |
| Supervised Baseline | 0.02 | 0.04 | 0.35 | 0.48 | 0.57 |
| Joint training | 0.51 | 0.76 | 0.93 | 0.97 | — |
| Supervised Pretraining | 0.55 | 0.48 | 0.68 | 0.96 | — |
| Unsupervised Pretraining | 0.75 | 0.81 | 0.90 | 0.95 | — |
| **CFQ** | $\eta = 0.01$ | $\eta = 0.02$ | $\eta = 0.04$ | $\eta = 0.08$ | $\eta = 0.99$ |
| Supervised Baseline | 0.11 | 0.23 | 0.36 | 0.48 | 0.53 |
| Joint training | 0.29 | 0.36 | 0.50 | 0.61 | — |
| Supervised Pretraining | 0.36 | 0.44 | 0.54 | 0.62 | — |
| Unsupervised Pretraining | 0.40 | 0.51 | 0.52 | 0.60 | — |

*Table 7.* Softmax DB – Teacher-forced X TA

| SCAN | $\eta = 0.04$ | $\eta = 0.08$ | $\eta = 0.16$ | $\eta = 0.32$ | $\eta = 0.99$ |
|---|---|---|---|---|---|
| Supervised Baseline | 0.66 | 0.77 | 0.84 | 0.88 | 0.88 |
| Joint training | 0.57 | 0.66 | 0.78 | 0.84 | — |
| Supervised Pretraining | 0.39 | 0.58 | 0.70 | 0.82 | — |
| Unsupervised Pretraining | 0.50 | 0.45 | 0.69 | 0.81 | — |
| PCFG | $\eta = 0.04$ | $\eta = 0.08$ | $\eta = 0.16$ | $\eta = 0.32$ | $\eta = 0.99$ |
| Supervised Baseline | 0.41 | 0.50 | 0.53 | 0.57 | 0.65 |
| Joint training | 0.50 | 0.54 | 0.57 | 0.61 | — |
| Supervised Pretraining | 0.47 | 0.50 | 0.54 | 0.57 | — |
| Unsupervised Pretraining | 0.48 | 0.50 | 0.61 | 0.63 | — |
| COGS | $\eta = 0.02$ | $\eta = 0.04$ | $\eta = 0.08$ | $\eta = 0.16$ | $\eta = 0.99$ |
| Supervised Baseline | 0.87 | 0.95 | 0.98 | 0.99 | 1.00 |
| Joint training | 0.90 | 0.96 | 0.99 | 1.00 | — |
| Supervised Pretraining | 0.00 | 0.88 | 0.95 | 0.99 | — |
| Unsupervised Pretraining | 0.88 | 0.96 | 0.98 | 0.99 | — |
| CFQ | $\eta = 0.01$ | $\eta = 0.02$ | $\eta = 0.04$ | $\eta = 0.08$ | $\eta = 0.99$ |
| Supervised Baseline | 0.74 | 0.79 | 0.82 | 0.85 | 0.88 |
| Joint training | 0.77 | 0.80 | 0.83 | 0.85 | — |
| Supervised Pretraining | 0.73 | 0.80 | 0.83 | 0.85 | — |
| Unsupervised Pretraining | 0.71 | 0.78 | 0.82 | 0.84 | — |

*Table 8.* Gumbel DB – Autoregressive Z TA

| SCAN | $\eta = 0.04$ | $\eta = 0.08$ | $\eta = 0.16$ | $\eta = 0.32$ | $\eta = 0.99$ |
|---|---|---|---|---|---|
| Supervised Baseline | 0.75 | 0.89 | 0.95 | 0.97 | 0.97 |
| Joint training | 0.76 | 0.88 | 0.95 | 0.96 | — |
| Supervised Pretraining | 0.80 | 0.93 | 0.96 | 0.97 | — |
| Unsupervised Pretraining | 0.79 | 0.90 | 0.96 | 0.96 | — |
| PCFG | $\eta = 0.04$ | $\eta = 0.08$ | $\eta = 0.16$ | $\eta = 0.32$ | $\eta = 0.99$ |
| Supervised Baseline | 0.16 | 0.25 | 0.70 | 0.82 | 0.89 |
| Joint training | 0.25 | 0.66 | 0.81 | 0.86 | — |
| Supervised Pretraining | 0.44 | 0.62 | 0.85 | 0.88 | — |
| Unsupervised Pretraining | 0.56 | 0.64 | 0.81 | 0.87 | — |
| COGS | $\eta = 0.02$ | $\eta = 0.04$ | $\eta = 0.08$ | $\eta = 0.16$ | $\eta = 0.99$ |
| Supervised Baseline | 0.74 | 0.80 | 0.84 | 0.85 | 0.86 |
| Joint training | 0.76 | 0.79 | 0.82 | 0.84 | — |
| Supervised Pretraining | 0.78 | 0.81 | 0.84 | 0.84 | — |
| Unsupervised Pretraining | 0.75 | 0.81 | 0.83 | 0.84 | — |
| CFQ | $\eta = 0.01$ | $\eta = 0.02$ | $\eta = 0.04$ | $\eta = 0.08$ | $\eta = 0.99$ |
| Supervised Baseline | 0.52 | 0.57 | 0.63 | 0.65 | 0.65 |
| Joint training | 0.56 | 0.62 | 0.65 | 0.69 | — |
| Supervised Pretraining | 0.55 | 0.61 | 0.65 | 0.68 | — |
| Unsupervised Pretraining | 0.60 | 0.66 | 0.68 | 0.71 | — |

*Table 9.* Gumbel DB – Autoregressive Z SA

| SCAN | $\eta = 0.04$ | $\eta = 0.08$ | $\eta = 0.16$ | $\eta = 0.32$ | $\eta = 0.99$ |
|---|---|---|---|---|---|
| Supervised Baseline | 0.13 | 0.46 | 0.79 | 0.86 | 0.89 |
| Joint training | 0.22 | 0.48 | 0.78 | 0.85 | — |
| Supervised Pretraining | 0.23 | 0.56 | 0.80 | 0.86 | — |
| Unsupervised Pretraining | 0.21 | 0.55 | 0.81 | 0.86 | — |
| PCFG | $\eta = 0.04$ | $\eta = 0.08$ | $\eta = 0.16$ | $\eta = 0.32$ | $\eta = 0.99$ |
| Supervised Baseline | 0.01 | 0.08 | 0.62 | 0.76 | 0.84 |
| Joint training | 0.06 | 0.57 | 0.74 | 0.80 | — |
| Supervised Pretraining | 0.26 | 0.51 | 0.78 | 0.81 | — |
| Unsupervised Pretraining | 0.45 | 0.55 | 0.75 | 0.81 | — |
| COGS | $\eta = 0.02$ | $\eta = 0.04$ | $\eta = 0.08$ | $\eta = 0.16$ | $\eta = 0.99$ |
| Supervised Baseline | 0.30 | 0.48 | 0.62 | 0.70 | 0.73 |
| Joint training | 0.41 | 0.56 | 0.64 | 0.68 | — |
| Supervised Pretraining | 0.38 | 0.53 | 0.63 | 0.66 | — |
| Unsupervised Pretraining | 0.32 | 0.53 | 0.63 | 0.66 | — |
| CFQ | $\eta = 0.01$ | $\eta = 0.02$ | $\eta = 0.04$ | $\eta = 0.08$ | $\eta = 0.99$ |
| Supervised Baseline | 0.14 | 0.23 | 0.30 | 0.33 | 0.34 |
| Joint training | 0.21 | 0.29 | 0.34 | 0.40 | — |
| Supervised Pretraining | 0.21 | 0.29 | 0.35 | 0.39 | — |
| Unsupervised Pretraining | 0.29 | 0.37 | 0.41 | 0.45 | — |

*Table 10.* Gumbel DB – Reconstruction Z TA

| SCAN | $\eta = 0.04$ | $\eta = 0.08$ | $\eta = 0.16$ | $\eta = 0.32$ | $\eta = 0.99$ |
|---|---|---|---|---|---|
| Supervised Baseline | 0.74 | 0.78 | 0.86 | 0.90 | 0.94 |
| Joint training | 0.96 | 0.94 | 0.96 | 0.95 | — |
| Supervised Pretraining | 0.97 | 0.98 | 0.97 | 0.97 | — |
| Unsupervised Pretraining | 0.81 | 0.88 | 0.90 | 0.94 | — |
| **PCFG** | $\eta = 0.04$ | $\eta = 0.08$ | $\eta = 0.16$ | $\eta = 0.32$ | $\eta = 0.99$ |
| Supervised Baseline | 0.33 | 0.46 | 0.70 | 0.75 | 0.79 |
| Joint training | 0.32 | 0.58 | 0.73 | 0.83 | — |
| Supervised Pretraining | 0.56 | 0.63 | 0.72 | 0.81 | — |
| Unsupervised Pretraining | 0.57 | 0.75 | 0.82 | 0.85 | — |
| **COGS** | $\eta = 0.02$ | $\eta = 0.04$ | $\eta = 0.08$ | $\eta = 0.16$ | $\eta = 0.99$ |
| Supervised Baseline | 0.90 | 0.93 | 0.96 | 0.97 | 0.98 |
| Joint training | 0.96 | 0.98 | 0.98 | 0.99 | — |
| Supervised Pretraining | 0.96 | 0.97 | 0.98 | 0.99 | — |
| Unsupervised Pretraining | 0.97 | 0.98 | 0.99 | 0.99 | — |
| **CFQ** | $\eta = 0.01$ | $\eta = 0.02$ | $\eta = 0.04$ | $\eta = 0.08$ | $\eta = 0.99$ |
| Supervised Baseline | 0.92 | 0.94 | 0.95 | 0.96 | 0.97 |
| Joint training | 0.95 | 0.96 | 0.96 | 0.97 | — |
| Supervised Pretraining | 0.94 | 0.95 | 0.96 | 0.97 | — |
| Unsupervised Pretraining | 0.98 | 0.98 | 0.98 | 0.98 | — |

*Table 11.* Gumbel DB – Reconstruction Z SA

| SCAN | $\eta = 0.04$ | $\eta = 0.08$ | $\eta = 0.16$ | $\eta = 0.32$ | $\eta = 0.99$ |
|---|---|---|---|---|---|
| Supervised Baseline | 0.05 | 0.08 | 0.17 | 0.27 | 0.30 |
| Joint training | 0.60 | 0.43 | 0.54 | 0.54 | — |
| Supervised Pretraining | 0.65 | 0.71 | 0.68 | 0.62 | — |
| Unsupervised Pretraining | 0.11 | 0.18 | 0.39 | 0.55 | — |
| **PCFG** | $\eta = 0.04$ | $\eta = 0.08$ | $\eta = 0.16$ | $\eta = 0.32$ | $\eta = 0.99$ |
| Supervised Baseline | 0.01 | 0.06 | 0.20 | 0.25 | 0.21 |
| Joint training | 0.01 | 0.11 | 0.19 | 0.38 | — |
| Supervised Pretraining | 0.07 | 0.11 | 0.15 | 0.25 | — |
| Unsupervised Pretraining | 0.05 | 0.21 | 0.32 | 0.42 | — |
| **COGS** | $\eta = 0.02$ | $\eta = 0.04$ | $\eta = 0.08$ | $\eta = 0.16$ | $\eta = 0.99$ |
| Supervised Baseline | 0.02 | 0.02 | 0.20 | 0.27 | 0.34 |
| Joint training | 0.29 | 0.48 | 0.59 | 0.61 | — |
| Supervised Pretraining | 0.24 | 0.42 | 0.56 | 0.60 | — |
| Unsupervised Pretraining | 0.30 | 0.46 | 0.59 | 0.63 | — |
| **CFQ** | $\eta = 0.01$ | $\eta = 0.02$ | $\eta = 0.04$ | $\eta = 0.08$ | $\eta = 0.99$ |
| Supervised Baseline | 0.05 | 0.10 | 0.17 | 0.21 | 0.21 |
| Joint training | 0.10 | 0.14 | 0.21 | 0.25 | — |
| Supervised Pretraining | 0.07 | 0.13 | 0.19 | 0.27 | — |
| Unsupervised Pretraining | 0.26 | 0.27 | 0.31 | 0.34 | — |

*Table 12.* Gumbel DB – Teacher-forced X TA

| SCAN | $\eta = 0.04$ | $\eta = 0.08$ | $\eta = 0.16$ | $\eta = 0.32$ | $\eta = 0.99$ |
|---|---|---|---|---|---|
| Supervised Baseline | 0.66 | 0.76 | 0.84 | 0.87 | 0.88 |
| Joint training | 0.60 | 0.70 | 0.78 | 0.85 | — |
| Supervised Pretraining | 0.40 | 0.63 | 0.76 | 0.84 | — |
| Unsupervised Pretraining | 0.36 | 0.64 | 0.62 | 0.76 | — |
| **PCFG** | $\eta = 0.04$ | $\eta = 0.08$ | $\eta = 0.16$ | $\eta = 0.32$ | $\eta = 0.99$ |
| Supervised Baseline | 0.39 | 0.48 | 0.51 | 0.55 | 0.57 |
| Joint training | 0.38 | 0.49 | 0.55 | 0.58 | — |
| Supervised Pretraining | 0.45 | 0.50 | 0.52 | 0.56 | — |
| Unsupervised Pretraining | 0.43 | 0.53 | 0.55 | 0.57 | — |
| **COGS** | $\eta = 0.02$ | $\eta = 0.04$ | $\eta = 0.08$ | $\eta = 0.16$ | $\eta = 0.99$ |
| Supervised Baseline | 0.84 | 0.93 | 0.97 | 0.98 | 0.99 |
| Joint training | 0.88 | 0.95 | 0.98 | 0.99 | — |
| Supervised Pretraining | 0.86 | 0.93 | 0.97 | 0.98 | — |
| Unsupervised Pretraining | 0.85 | 0.94 | 0.98 | 0.99 | — |
| **CFQ** | $\eta = 0.01$ | $\eta = 0.02$ | $\eta = 0.04$ | $\eta = 0.08$ | $\eta = 0.99$ |
| Supervised Baseline | 0.71 | 0.77 | 0.80 | 0.83 | 0.85 |
| Joint training | 0.75 | 0.79 | 0.81 | 0.84 | — |
| Supervised Pretraining | 0.72 | 0.78 | 0.81 | 0.84 | — |
| Unsupervised Pretraining | 0.69 | 0.77 | 0.81 | 0.84 | — |

*Table 13.* VQ DB – Autoregressive Z TA

| SCAN | $\eta = 0.04$ | $\eta = 0.08$ | $\eta = 0.16$ | $\eta = 0.32$ | $\eta = 0.99$ |
|---|---|---|---|---|---|
| Supervised Baseline | 0.73 | 0.89 | 0.98 | 1.00 | 1.00 |
| Joint training | 0.67 | 0.88 | 0.95 | 0.97 | — |
| Supervised Pretraining | 0.84 | 0.95 | 0.99 | 1.00 | — |
| Unsupervised Pretraining | 0.75 | 0.91 | 0.97 | 0.99 | — |
| PCFG | $\eta = 0.04$ | $\eta = 0.08$ | $\eta = 0.16$ | $\eta = 0.32$ | $\eta = 0.99$ |
| Supervised Baseline | 0.23 | 0.86 | 0.87 | 0.93 | 0.93 |
| Joint training | 0.31 | 0.66 | 0.90 | 0.89 | — |
| Supervised Pretraining | 0.41 | 0.86 | 0.90 | 0.93 | — |
| Unsupervised Pretraining | 0.12 | 0.13 | 0.13 | 0.13 | — |
| COGS | $\eta = 0.02$ | $\eta = 0.04$ | $\eta = 0.08$ | $\eta = 0.16$ | $\eta = 0.99$ |
| Supervised Baseline | 0.86 | 0.95 | 0.99 | 1.00 | 1.00 |
| Joint training | 0.92 | 0.97 | 0.99 | 0.99 | — |
| Supervised Pretraining | 0.91 | 0.97 | 0.99 | 1.00 | — |
| Unsupervised Pretraining | 0.94 | 0.97 | 0.98 | 0.98 | — |
| CFQ | $\eta = 0.01$ | $\eta = 0.02$ | $\eta = 0.04$ | $\eta = 0.08$ | $\eta = 0.99$ |
| Supervised Baseline | 0.49 | 0.77 | 0.91 | 0.95 | 0.84 |
| Joint training | 0.71 | 0.81 | 0.91 | 0.96 | — |
| Supervised Pretraining | 0.69 | 0.83 | 0.94 | 0.96 | — |
| Unsupervised Pretraining | 0.44 | 0.54 | 0.52 | 0.61 | — |

*Table 14.* VQ DB – Autoregressive Z SA

| SCAN | $\eta = 0.04$ | $\eta = 0.08$ | $\eta = 0.16$ | $\eta = 0.32$ | $\eta = 0.99$ |
|---|---|---|---|---|---|
| Supervised Baseline | 0.10 | 0.41 | 0.85 | 0.98 | 1.00 |
| Joint training | 0.12 | 0.44 | 0.70 | 0.87 | — |
| Supervised Pretraining | 0.29 | 0.64 | 0.91 | 0.98 | — |
| Unsupervised Pretraining | 0.15 | 0.54 | 0.84 | 0.93 | — |
| PCFG | $\eta = 0.04$ | $\eta = 0.08$ | $\eta = 0.16$ | $\eta = 0.32$ | $\eta = 0.99$ |
| Supervised Baseline | 0.04 | 0.86 | 0.87 | 0.94 | 0.91 |
| Joint training | 0.14 | 0.59 | 0.91 | 0.89 | — |
| Supervised Pretraining | 0.23 | 0.86 | 0.90 | 0.94 | — |
| Unsupervised Pretraining | 0.00 | 0.00 | 0.00 | 0.00 | — |
| COGS | $\eta = 0.02$ | $\eta = 0.04$ | $\eta = 0.08$ | $\eta = 0.16$ | $\eta = 0.99$ |
| Supervised Baseline | 0.47 | 0.74 | 0.92 | 0.97 | 0.84 |
| Joint training | 0.51 | 0.81 | 0.93 | 0.97 | — |
| Supervised Pretraining | 0.59 | 0.82 | 0.94 | 0.97 | — |
| Unsupervised Pretraining | 0.51 | 0.63 | 0.62 | 0.59 | — |
| CFQ | $\eta = 0.01$ | $\eta = 0.02$ | $\eta = 0.04$ | $\eta = 0.08$ | $\eta = 0.99$ |
| Supervised Baseline | 0.01 | 0.53 | 0.79 | 0.87 | 0.25 |
| Joint training | 0.69 | 0.78 | 0.92 | 0.96 | — |
| Supervised Pretraining | 0.41 | 0.66 | 0.85 | 0.91 | — |
| Unsupervised Pretraining | 0.01 | 0.02 | 0.07 | 0.05 | — |

*Table 15.* VQ DB – Reconstruction Z TA

| SCAN | $\eta = 0.04$ | $\eta = 0.08$ | $\eta = 0.16$ | $\eta = 0.32$ | $\eta = 0.99$ |
|---|---|---|---|---|---|
| Supervised Baseline | 0.73 | 0.77 | 0.83 | 0.91 | 0.99 |
| Joint training | 0.95 | 0.95 | 0.95 | 0.96 | — |
| Supervised Pretraining | 0.99 | 0.98 | 0.98 | 0.97 | — |
| Unsupervised Pretraining | 0.86 | 0.97 | 0.99 | 0.99 | — |
| PCFG | $\eta = 0.04$ | $\eta = 0.08$ | $\eta = 0.16$ | $\eta = 0.32$ | $\eta = 0.99$ |
| Supervised Baseline | 0.35 | 0.67 | 0.77 | 0.81 | 0.53 |
| Joint training | 0.59 | 0.77 | 0.91 | 0.93 | — |
| Supervised Pretraining | 0.68 | 0.81 | 0.83 | 0.89 | — |
| Unsupervised Pretraining | 0.32 | 0.38 | 0.32 | 0.32 | — |
| COGS | $\eta = 0.02$ | $\eta = 0.04$ | $\eta = 0.08$ | $\eta = 0.16$ | $\eta = 0.99$ |
| Supervised Baseline | 0.93 | 0.98 | 0.99 | 1.00 | 0.98 |
| Joint training | 0.97 | 0.99 | 0.99 | 1.00 | — |
| Supervised Pretraining | 0.97 | 0.98 | 0.99 | 1.00 | — |
| Unsupervised Pretraining | 0.98 | 0.99 | 0.98 | 0.98 | — |
| CFQ | $\eta = 0.01$ | $\eta = 0.02$ | $\eta = 0.04$ | $\eta = 0.08$ | $\eta = 0.99$ |
| Supervised Baseline | 0.89 | 0.95 | 0.98 | 0.98 | 0.98 |
| Joint training | 0.97 | 0.98 | 0.99 | 0.99 | — |
| Supervised Pretraining | 0.97 | 0.98 | 0.99 | 0.99 | — |
| Unsupervised Pretraining | 0.94 | 0.94 | 0.94 | 0.94 | — |

*Table 16.* VQ DB – Reconstruction Z SA

| SCAN | $\eta = 0.04$ | $\eta = 0.08$ | $\eta = 0.16$ | $\eta = 0.32$ | $\eta = 0.99$ |
|---|---|---|---|---|---|
| Supervised Baseline | 0.05 | 0.12 | 0.29 | 0.61 | 0.91 |
| Joint training | 0.40 | 0.38 | 0.39 | 0.52 | — |
| Supervised Pretraining | 0.89 | 0.76 | 0.71 | 0.66 | — |
| Unsupervised Pretraining | 0.17 | 0.67 | 0.90 | 0.85 | — |
| PCFG | $\eta = 0.04$ | $\eta = 0.08$ | $\eta = 0.16$ | $\eta = 0.32$ | $\eta = 0.99$ |
| Supervised Baseline | 0.02 | 0.18 | 0.19 | 0.25 | 0.12 |
| Joint training | 0.10 | 0.30 | 0.64 | 0.75 | — |
| Supervised Pretraining | 0.13 | 0.33 | 0.39 | 0.56 | — |
| Unsupervised Pretraining | 0.00 | 0.00 | 0.00 | 0.00 | — |
| COGS | $\eta = 0.02$ | $\eta = 0.04$ | $\eta = 0.08$ | $\eta = 0.16$ | $\eta = 0.99$ |
| Supervised Baseline | 0.32 | 0.68 | 0.78 | 0.95 | 0.39 |
| Joint training | 0.41 | 0.68 | 0.86 | 0.93 | — |
| Supervised Pretraining | 0.35 | 0.69 | 0.88 | 0.96 | — |
| Unsupervised Pretraining | 0.44 | 0.57 | 0.52 | 0.53 | — |
| CFQ | $\eta = 0.01$ | $\eta = 0.02$ | $\eta = 0.04$ | $\eta = 0.08$ | $\eta = 0.99$ |
| Supervised Baseline | 0.01 | 0.19 | 0.51 | 0.52 | 0.22 |
| Joint training | 0.27 | 0.40 | 0.57 | 0.67 | — |
| Supervised Pretraining | 0.19 | 0.41 | 0.56 | 0.71 | — |
| Unsupervised Pretraining | 0.00 | 0.01 | 0.00 | 0.00 | — |

*Table 17.* VQ DB – Teacher-forced X TA

| SCAN | $\eta = 0.04$ | $\eta = 0.08$ | $\eta = 0.16$ | $\eta = 0.32$ | $\eta = 0.99$ |
|---|---|---|---|---|---|
| Supervised Baseline | 0.65 | 0.73 | 0.81 | 0.86 | 0.88 |
| Joint training | 0.56 | 0.67 | 0.74 | 0.84 | — |
| Supervised Pretraining | 0.41 | 0.62 | 0.71 | 0.78 | — |
| Unsupervised Pretraining | 0.27 | 0.61 | 0.57 | 0.79 | — |
| **PCFG** | $\eta = 0.04$ | $\eta = 0.08$ | $\eta = 0.16$ | $\eta = 0.32$ | $\eta = 0.99$ |
| Supervised Baseline | 0.32 | 0.52 | 0.55 | 0.58 | 0.62 |
| Joint training | 0.43 | 0.54 | 0.59 | 0.65 | — |
| Supervised Pretraining | 0.48 | 0.53 | 0.55 | 0.58 | — |
| Unsupervised Pretraining | 0.33 | 0.34 | 0.33 | 0.33 | — |
| **COGS** | $\eta = 0.02$ | $\eta = 0.04$ | $\eta = 0.08$ | $\eta = 0.16$ | $\eta = 0.99$ |
| Supervised Baseline | 0.85 | 0.96 | 0.99 | 0.99 | 0.97 |
| Joint training | 0.88 | 0.95 | 0.98 | 0.99 | — |
| Supervised Pretraining | 0.85 | 0.93 | 0.98 | 0.99 | — |
| Unsupervised Pretraining | 0.88 | 0.93 | 0.93 | 0.93 | — |
| **CFQ** | $\eta = 0.01$ | $\eta = 0.02$ | $\eta = 0.04$ | $\eta = 0.08$ | $\eta = 0.99$ |
| Supervised Baseline | 0.60 | 0.77 | 0.83 | 0.86 | 0.88 |
| Joint training | 0.74 | 0.78 | 0.83 | 0.86 | — |
| Supervised Pretraining | 0.68 | 0.78 | 0.84 | 0.86 | — |
| Unsupervised Pretraining | 0.49 | 0.55 | 0.53 | 0.53 | — |